%% file: main.tex
\documentclass[runningheads]{llncs}

\usepackage{eccv}

\usepackage{eccvabbrv}

\usepackage{graphicx}
\usepackage{booktabs}
\usepackage[normalem]{ulem}
\useunder{\uline}{\ul}{}

\usepackage[accsupp]{axessibility}  

\usepackage[pagebackref,breaklinks,colorlinks]{hyperref}

\usepackage{orcidlink}

\input{preamble}

\begin{document}

\title{UV Gaussians: Joint Learning of Mesh Deformation and Gaussian Textures for Human Avatar Modeling}

\titlerunning{UV Gaussians}

\author{Yujiao Jiang\inst{1} \and
    Qingmin Liao\inst{1} \and
    Xiaoyu Li\inst{2}\thanks{Corresponding author} \and
    Li Ma\inst{3} \and
    Qi Zhang\inst{2} \and
    Chaopeng Zhang\inst{2} \and
    Zongqing Lu\inst{1} \and
    Ying Shan\inst{2}}

\authorrunning{Y. Jiang et al.}

\institute{Tsinghua Shenzhen International Graduate School, Tsinghua University \and
Tencent AI Lab\and
The Hong Kong University of Science and Technology
}

\maketitle

\begin{abstract}
  Reconstructing photo-realistic drivable human avatars from multi-view image sequences has been a popular and challenging topic in the field of computer vision and graphics. While existing NeRF-based methods can achieve high-quality novel view rendering of human models, both training and inference processes are time-consuming. Recent approaches have utilized 3D Gaussians to represent the human body, enabling faster training and rendering.
  However, they undermine the importance of the mesh guidance and directly predict Gaussians in 3D space with coarse mesh guidance. This hinders the learning procedure of the Gaussians and tends to produce blurry textures. 
  Therefore, we propose UV Gaussians, which models the 3D human body by jointly learning mesh deformations and 2D UV-space Gaussian textures. 
  We utilize the embedding of UV map to learn Gaussian textures in 2D space, leveraging the capabilities of powerful 2D networks to extract features. Additionally, through an independent Mesh network, we optimize pose-dependent geometric deformations, thereby guiding Gaussian rendering and significantly enhancing rendering quality. We collect and process a new dataset of human motion, which includes multi-view images, scanned models, parametric model registration, and corresponding texture maps. Experimental results demonstrate that our method achieves state-of-the-art synthesis of novel view and novel pose. The code and data will be made available on the homepage \url{https://alex-jyj.github.io/UV-Gaussians/} once the paper is accepted.
\end{abstract}

\input{sec/1_introduction}
\input{sec/2_related_work}
\input{sec/3_method}
\input{sec/4_experiments}

\input{sec/5_conclusion}

%
%
\bibliographystyle{splncs04}
\bibliography{egbib}

\clearpage
\appendix

\input{sec/6_suppl}
\end{document}

%% file: preamble.tex
\newcommand{\col}{{\mathbf{c}}}

\newcommand{\cov}{\Sigma}
\newcommand{\covb}{\Sigma'}
\newcommand{\mean}{\boldsymbol\mu}
\newcommand{\scale}{\boldsymbol\sigma}
\newcommand{\rot}{\mathbf{q}}
\newcommand{\opacity}{\alpha}

%% file: sec/1_introduction.tex
\section{Introduction}
\label{sec:intro}

\begin{figure}[tb]
	\centering
	\includegraphics[width=1.0\columnwidth]{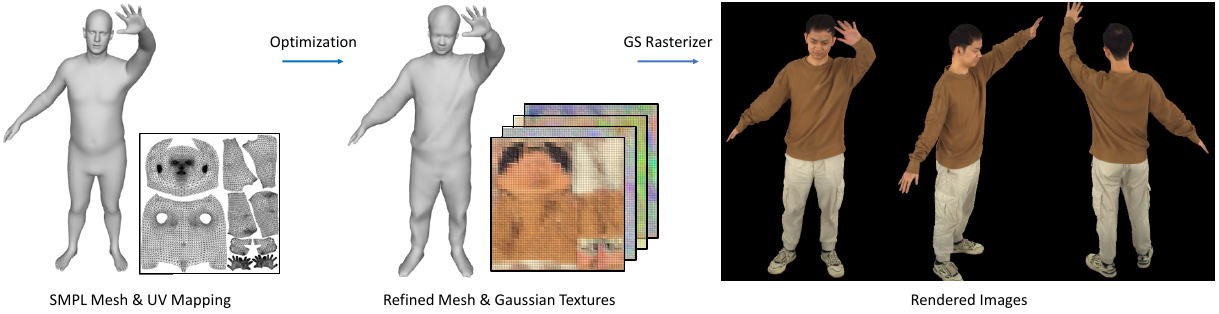}
	\caption{Based on the SMPL mesh and its UV mapping, we learn pose-dependent refined mesh and its Gaussian textures. By combining the advantages of high-quality rendering from Gaussian Splatting and easy animation of template mesh, our method could produce photo-realistic human avatars.}
        \label{fig:teaser}
\end{figure}

Reconstructing high-fidelity and photo-realistic human avatars has long been a significant topic in the fields of computer vision and computer graphics, with various applications in games, movies, and virtual/augmented reality. Traditional human modeling requires a time-consuming and tedious pipeline involving modeling, rigging, and skinning, yet it remains challenging to create photo-realistic avatars.


Early works~\cite{kanazawa2018end, kolotouros2019learning} often employ explicit mesh representations to depict 3D human bodies, commonly utilizing parametric mesh models such as SMPL~\cite{SMPL:2015} and SMPL-X~\cite{SMPL-X:2019}. 
Typically, these approaches involve regressing low-dimensional model parameters to accurately align model projections with the images.
Owing to the low-dimensional nature of the parametric space, these methods are incapable of capturing clothing wrinkles and detailed textures. Subsequent efforts~\cite{alldieck2018video, alldieck2018detailed, alldieck2019learning, bhatnagar2019multi, bhatnagar2020combining, ma2020learning} have extended parametric models by incorporating vertex displacements to represent clothing, improving the ability to express geometric variations. However, they are still constrained by the fixed topological structure of parametric models, making it difficult to represent complex and detailed structures. 

With the development of implicit representations, an increasing number of studies have started to introduce these representations for 3D human modeling, such as Signed Distance Functions (SDF)~\cite{park2019deepsdf} and Neural Radiance Fields (NeRF)~\cite{mildenhall2021nerf}. Neural Body first applies the NeRF representation to human modeling, demonstrating promising results at that time. Subsequently, a significant amount of work~\cite{peng2021neural, peng2021animatable, wang2022arah, ho2023learning, weng2022humannerf, zheng2022structured, chen2023uv, li2023posevocab} has been explored to utilize NeRF representation for human modeling. Many of them learn to transform the points in observation space to canonical space and establish NeRF in canonical space to achieve animation. However, the NeRF-based methods, while achieving high-quality results, are computationally intensive and time-consuming due to the requirements for intensive inference of the MLP network, thus constraining their usage. 

Recently, the emergence of 3D Gaussian Splatting (3DGS)~\cite{kerbl20233d} has attracted significant attention. Building upon traditional point-based rendering methods, 3DGS has significantly enhanced rendering quality and enables real-time rendering. Researchers have started exploring the use of 3D Gaussians to represent 3D human avatars. Approaches such as 3DGS-Avatar~\cite{qian20233dgs} and GauHuman~\cite{hu2023gauhuman} initialize Gaussians using points sampled from the SMPL model, then employ supervised learning on monocular image sequences to optimize the Gaussian points' features, striking a promising balance between real-time performance and rendering quality. Concurrently, the initiatives of Animatable Gaussians~\cite{li2023animatable}, GaussianAvatar~\cite{hu2023gaussianavatar}, and ASH~\cite{pang2023ash} have ventured into learning Gaussian features from 2D images, allowing for the leveraging of powerful 2D networks to model realistic human models. Animatable Gaussians, for instance, utilizes two position maps (front and back views) as network inputs but overlooks the unique characteristics of facial and hand regions, which, although comprising a few pixels, entail intricate textures. In contrast, GaussianAvatar employs the SMPL model as the guidance which may fail to model detailed structure movements effectively.

Hence, we propose UV Gaussians, which amalgamate the effortless animation of parametric mesh models with the high-quality rendering of 3DGS. Instead of directly using the SMPL mesh as the basis, we adopt a template mesh with personalized geometry and learn a pose-dependent displacement to deform this template, thereby modeling pose-dependent variations. With this refined mesh as the guidance, 3DGS could achieve better results. Simultaneously, we introduce a Gaussian U-Net that takes the position maps of the mesh and an average texture map as input to learn the features of Gaussian points in UV space, referred to as Gaussian Textures in this paper. The learned Gaussians are then rendered guided by the refined mesh, achieving high-quality and photo-realistic renderings as shown in Fig.~\ref{fig:teaser}. Additionally, we collect and process a high-quality human motion dataset comprising multi-view image sequences, scan models, SMPL-X registration, SMPLX-D fitting, and corresponding texture maps. Through experiments on multiple human data, our algorithm demonstrates state-of-the-art effectiveness in novel view and novel pose synthesis, validating the efficacy of our algorithm. 

In summary, our contributions in this work include: 
\begin{itemize}
  \item[$\bullet$] We propose UV Gaussians, a novel method for human modeling combining 3DGS expressed in UV space with mesh optimization to achieve photo-realistic drivable human model rendering. 
  \item[$\bullet$] A specialized Mesh U-Net designed for geometric optimization has been developed to produce a refined mesh output to guide 3DGS animation.
  \item[$\bullet$] A high-quality human motion dataset encompassing multi-view images, scan models, and fitted parameterized models with textures.
\end{itemize}

%% file: sec/2_related_work.tex
\section{Related Work}

\subsection{Mesh-based Human Avatars}
Human avatars are traditionally represented as explicit polygon meshes, a representation that is particularly well-suited for conventional graphics rendering pipelines. Some studies~\cite{alldieck2018video, alldieck2019learning, bhatnagar2020combining, bhatnagar2020loopreg} reconstruct the human body mesh by optimizing the mesh deformation based on parametric body models such as SMPL~\cite{SMPL:2015}. These methods can obtain the mesh with the same topology as the parametric models, making them compatible with these models and easily driven by their pose parameters. However, the constrained topology structure of parametric models limits their ability to represent complex clothing effectively. To address this, another line of research~\cite{guan2012drape, lahner2018deepwrinkles, gundogdu2019garnet, patel2020tailornet, santesteban2021self} focuses on modeling clothing as meshes separate from the human body, aiming to generate accurate and realistic clothing deformation. Nonetheless, it remains challenging to model and animate various types of clothing using a single pre-defined template mesh. In this work, we learn pose-dependent vertex displacements of a topology-consistent human mesh with clothing, which enables us to model pose-dependent variation while maintaining compatibility with parametric body models for animation. With the guidance of this mesh representation, 3D Gaussians can achieve accurate initial geometry and corresponding movement for animation.

\subsection{NeRF-based Human Avatars}
Recently, implicit representations such as NeRF~\cite{mildenhall2021nerf}, occupancy~\cite{mescheder2019occupancy}, and SDF~\cite{park2019deepsdf} have demonstrated remarkable results due to their ability to model arbitrary topologies. As a result, these representations have been also adopted for modeling clothed humans~\cite{noguchi2021neural, peng2021neural, peng2021animatable, su2021nerf, xu2021h, wang2022arah, ho2023learning, feng2022capturing, jiang2022neuman, weng2022humannerf, liu2021neural, chen2023uv, zheng2022structured, li2023posevocab}. Among these works, Neural Body~\cite{peng2021neural} introduces an implicit neural representation that is conditioned with structured latent codes anchored to the mesh vertices. While this method can generate photorealistic novel views from sparse multi-view videos, it struggles to generalize effectively to novel poses. To map posed space into a canonical pose, Animatable NeRF~\cite{peng2021animatable} learns neural blend weight fields and Neural Actor~\cite{liu2021neural} utilizes a coarse body model as a proxy. In addition, UV Volumes~\cite{chen2023uv} represents the dynamic human as a 3D UV volume and 2D neural texture for high-fidelity rendering and editing. ARAH~\cite{wang2022arah} utilizes SDF representation and a root-finding algorithm to map each point to canonical space through forward linear blend skinning. By doing so, it facilitates the creation of avatars in extreme out-of-distribution poses. Recently, SLRF~\cite{zheng2022structured} employs structured local NeRFs attached to pre-defined nodes on the SMPL model, while PoseVocab~\cite{li2023posevocab} interpolates pose embeddings at various spatial levels to obtain the conditional pose feature, both of which have achieved impressive results. In contrast, our approach adopts 3D Gaussian Splatting~\cite{kerbl20233d} that allows for both real-time and higher-quality rendering compared to NeRFs.

\subsection{3D Gaussian Splatting}
The NeRF-based representation has significantly transformed the field of novel view synthesis due to its compact formulation and high-quality rendering. However, the implicit neural networks adopted by NeRFs are costly to train and render. To address this issue, Instant NGP~\cite{muller2022instant} utilizes multiresolution hash encoding to accelerate training processes. 3D Gaussian Splatting~\cite{kerbl20233d} introduces an efficient differentiable point-based rendering method for high-quality and real-time rendering, which has recently garnered significant attention and has also been employed in human avatar modeling. Several concurrent works~\cite{hu2023gaussianavatar, hu2023gauhuman, kocabas2023hugs, jena2023splatarmor, qian20233dgs} extend 3D Gaussian Splatting to model human avatars from a single video. However, monocular videos offer only sparse observations of the scene, making it challenging to accurately reconstruct the Gaussian avatar, which typically necessitates multi-view images. Other concurrent works~\cite{zielonka2023drivable, li2023animatable} also build high-quality avatars based on multi-view inputs using Gaussian splatting. Specifically, D3GA~\cite{zielonka2023drivable} uses Tetrahedral cage-based deformations for 3D Gaussians. Animatable Gaussians~\cite{li2023animatable} parameterizes 3D Gaussians onto the front and back Gaussian maps and employs 2D CNNs to learn these maps. In this work, we use CNNs to learn Gaussian textures which parameterize 3D Gaussians in UV space that covers the whole surface to model pose-dependent variation. To animate 3D Gaussians more accurately, we also learn a mesh deformation based on a fitted template mesh, which guides the motion of the 3D Gaussians.

%% file: sec/3_method.tex
\section{Preliminary}
In this section, we briefly introduce some concepts related to Linear Blend Skinning and 3D Gaussian Splatting.

\subsection{Linear Blend Skinning}
Linear Blend Skinning (LBS) is a weight-based technique that associates each vertex with one or more bones and uses weight values to describe the influence of each bone on the vertex. Vertex deformation is calculated by linearly interpolating transformations on the associated bones: 
\begin{equation}
    \mathbf{p}^\prime = \sum_{j=1}^{J}w_j(\mathbf{p})T_j\mathbf{p} \text{,}
\end{equation}
where $J$ represents the number of joints, $N$ represents the number of vertices,
$\mathbf{p}^\prime \in \mathbb{R}^{N\times3}$ is the new position of the skinned vertex, $w \in \mathbb{R}^{N \times J}$ is the weight matrix, $T \in \mathbb{R}^{J\times 4\times 4}$ is the affine transformation matrix of each joint representing rotation and translation, and $\mathbf{p}\in \mathbb{R}^{N\times3}$ is the original mesh vertex position.

\subsection{3D Gaussian Splatting}
3D Gaussian Splatting \cite{kerbl20233d} represents a static 3D scene as a set of 3D isotropic Gaussian balls defined as 3D covariance matrix $\cov$ and 3D mean $\mean$. To render a 3D Gaussian, we splat the Gaussian into 2D by:
\begin{equation}
    \covb = J W \cov W^T J^T\text{,}
\end{equation}
where $W$ is the camera rotation matrix and $J$ is the affine approximation of the projective transformation. 
The covariance matrix $\cov$ is typically derived through a scale matrix $S$ and a rotation matrix $R$ to ensure it is positive semi-definite:
\begin{equation}
    \cov = R S S^T R^T\text{,}
\end{equation}
By taking the upper-left corner of $\covb$, we obtain the image-space Gaussian, which could later be rasterized into images. In the implementation, a 3D Gaussian is often parameterized using their position $\mean \in \mathbb{R}^3$, scale $\scale \in \mathbb{R}^3$, rotation $\rot \in \mathbb{H}$, opacity $\opacity \in \mathbb{R}$ and color $\col \in \mathbb{R}^3$, where $\scale$ is the diagonal vector of the scale matrix $S$, while $\rot$ represents the rotation matrix $R$ using a quaternion vector. To render a cloud of 3D Gaussians, we sort the $N$ Gaussians from far to near at certain view point, splat them sequentially, and accumulate the transmittance and the color to get the final image:
\begin{equation}
    C = \sum_{i=1}^{N} c_i\alpha_i^\prime \prod_{j=1}^{i-1}(1-\alpha_j^\prime) \text{,}
\end{equation}
where $c_i$ represents the color of each 2D Gaussian, while $\alpha_i^\prime$ denotes the probability density of the 2D Gaussian at the corresponding pixel position multiplied by the learned opacity $\alpha_i$.
Since this process is fully differentiable, we could obtain a gradient of the Gaussian parameters using rendering errors. 

\section{Method}
\begin{figure}[tb]
	\centering
	\includegraphics[width=1.0\columnwidth]{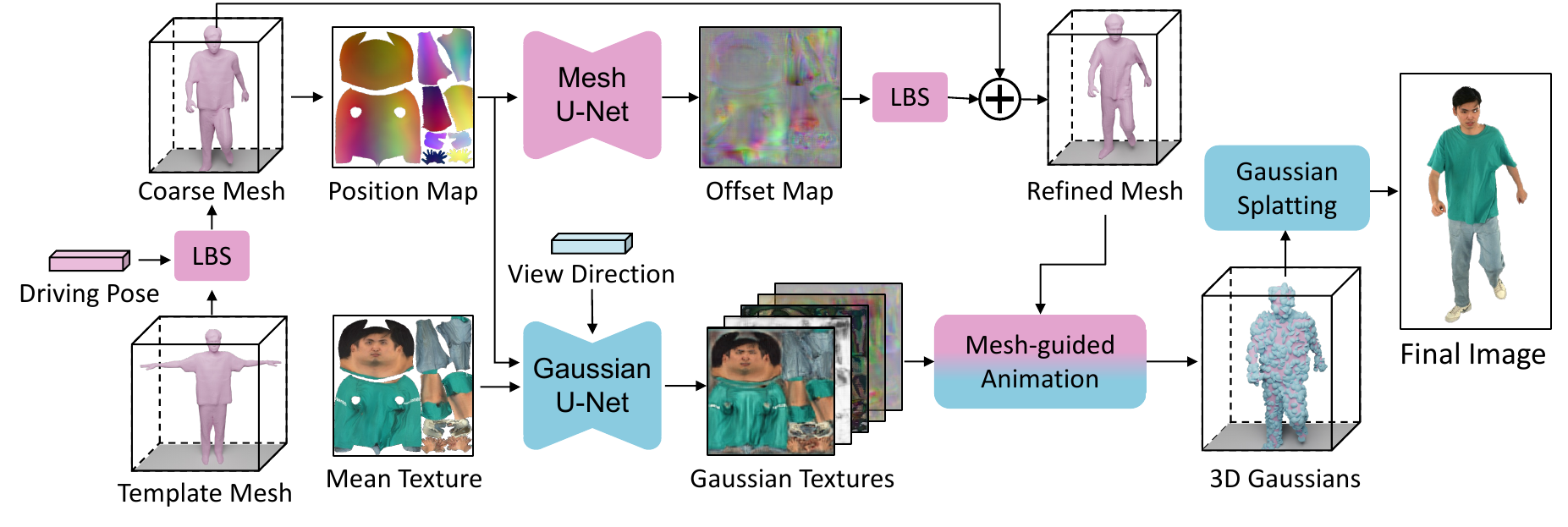}
	\caption{Overview of our method, which comprises three primary modules: a Mesh U-Net for learning pose-dependent mesh deformation, a Gaussian U-Net for learning pose-dependent Gaussian textures, and a Mesh-Guided 3D Gaussian Animation for animating the Gaussian guided by the mesh.}
        \label{fig:pipeline}
\end{figure}

Given a set of calibrated and synchronized multi-view videos of a dynamic human, the objective of our approach is to create a photorealistic, animatable avatar. The overview of our method, as depicted in Fig.~\ref{fig:pipeline}, comprises three primary modules: a Mesh U-Net that learns pose-dependent mesh deformation based on a posed template mesh (Sec.~\ref{sec:mesh_deformation}), a Gaussian U-Net that learns pose-dependent Gaussian textures which represent Gaussian attributes in UV texture space (Sec.~\ref{sec:gaussian_texture}), and a Mesh-guided 3D Gaussians Animation to guide the motion of Gaussians in 3D space by the refined mesh (Sec.~\ref{sec:animation}). The pre-processed data these modules rely on will be illustrated in Sec.~\ref{sec:data}.

\subsection{Data Processing}
\label{sec:data}
To fit the parameters of the SMPL-X model~\cite{SMPL-X:2019}, we first estimate the 2D keypoints of the multi-view images using Openpose~\cite{OpenPose}. This is followed by the estimation of 3D keypoints through triangulation, and subsequently fitting the SMPL-X model using EasyMocap~\cite{easymocap}. The SMPL mesh, however, can only represent the naked human body without clothing, providing only an approximation of the actual human body geometry. To improve the parametric mesh model's capacity to capture intricate geometric details while keeping topology consistency with SMPL-X for seamless animation, we initially reconstruct the scan mesh for each frame by employing MVS methods. Subsequently, we optimize the vertex displacement of the SMPL-X mesh to align it with the scanned model's mesh, leading to the SMPLX-D model. To tackle the challenge of fitting the complex facial area in the SMPLX-D model, we implement a targeted optimization method. Additionally, the texture of the fitted SMPLX-D model is baked to inherit the texture from the scanned model, while ensuring consistency in the UV mapping. We give an example of this processing flow and more details of the data processing in the supplementary materials~\ref{sec:supp-dataprocess}.


\subsection{Pose-dependent Mesh Deformation}
\label{sec:mesh_deformation}

Through the above processing flow, we could obtain the SMPLX-D mesh model that includes clothing geometry with its texture map for each captured frame. To learn a pose-dependent SMPLX-D mesh for novel poses beyond captured frames, we first select a frame with a pose close to a T-pose from the captured frames as a reference. We then deform this pose to a standard T-pose using Linear Blend Skinning (LBS). This deformed standard T-pose $\boldsymbol{G}_{T}$ will serve as the template mesh for all poses. Given a driving pose with pose parameters $\boldsymbol{\Theta}$, we deform the T-pose mesh $\boldsymbol{G}_{T}$ using Linear Blend Skinning based on the pose parameters $\boldsymbol{\Theta}$ and obtain the posed coarse mesh $\boldsymbol{G}_{coarse}$. Subsequently, a Mesh U-Net is introduced to learn the pose-dependent mesh deformation based on this coarse mesh. 

Previous approaches~\cite{peng2021neural, chen2023uv} discretize the 3D space into volumes and employ 3D convolutions to extract features and predict the geometry. However, this approach results in high computational costs and increasing difficulty in generalizing well to novel poses due to the complexity of operating in 3D space. To tackle these challenges, we adopt a strategy of rendering the mesh vertices in UV space and learning the 3D vertex offsets in this space, which also maintains topology consistency across different poses. Specifically, the 3D vertex coordinates of the mesh $\boldsymbol{G}_{course}$ are first rasterized into UV space to generate a position map $\mathcal{P}$, which serves as input to the mesh network $\mathcal{M}$, represented as Mesh U-Net. Leveraging a fundamental U-Net~\cite{ronneberger2015u} architecture for the mesh network, the network $\mathcal{M}$ predicts a vertex offset map $\mathcal{P}_{offset}$ based on the input position map:

\begin{equation}
    \mathcal{P}_{offset} = \mathcal{M}\left(\mathcal{P}(\boldsymbol{\Theta})\right) \text{.}
\end{equation}

The corresponding offset for each vertex is interpolated based on its UV coordinates from the offset map $\mathcal{P}_{offset}$, resulting in the mesh vertex offset $\boldsymbol{G}_{offset}$. This offset is added to the T-pose mesh $\boldsymbol{G}_{T}$, later transformed into the posed space using LBS. This process ultimately yields a refined mesh $\boldsymbol{G}_{refined}$ capable of capturing pose-dependent geometric alterations:

\begin{equation}
    \boldsymbol{G}_{refined} = LBS(\boldsymbol{G}_{T} + \boldsymbol{G}_{offset}, \boldsymbol{\Theta}) = \boldsymbol{G}_{coarse} + LBS(\boldsymbol{G}_{offset}, \boldsymbol{\Theta})\text{.} 
\end{equation}

\begin{figure}[tbp]
	\centering
	\includegraphics[width=1.0\columnwidth]{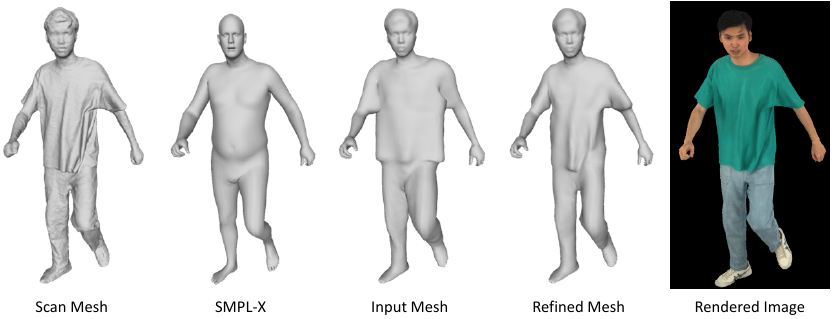}
	\caption{Example of mesh deformation. By optimizing the vertices offsets, our method could achieve a refined mesh with more accurate geometry while keeping the topology consistency with SMPL-X for animation. This refined mesh can be used to guide the rendering of 3D Gaussians, resulting in photorealistic results.}
        \label{fig:mesh}
\end{figure}

The refined mesh $\boldsymbol{G}_{refined}$ will later serve as an internal guide mesh, facilitating the animation of 3D Gaussians as shown in Fig.~\ref{fig:mesh}.

\subsection{Pose-dependent Gaussian Textures}
\label{sec:gaussian_texture}
Recent research~\cite{li2023animatable} has been explored to leverage powerful 2D CNNs to learn 3D Gaussians. Animatable Gaussians~\cite{li2023animatable} parameterizes 3D Gaussians onto the front and back Gaussian maps and takes the mesh's front and back side position maps as input of a 2D convolutional network to learn these maps, which enhances computational efficiency. Nevertheless, it is noted that this strategy overlooks regions like the head and hands, which, despite occupying a small pixel size, possess intricate geometric and textural features. Furthermore, it neglects the region in the side view, which is not covered by the front and back views. To tackle this issue, we parameterize 3D Gaussians in UV space, referred to as Gaussian Textures in this paper, which can project each pixel to a 3D Gaussian through UV mapping. This modification enables our approach to accurately depict details in small regions and the whole geometry.

In the implementation of the Gaussian network $\mathcal{G}$, we take the averaged texture map $\mathcal{T}$ from all the poses in captured frames as input to provide initial color information for 3D Gaussians. Additionally, we provide the position map $\mathcal{P}$ to the network to supply pixel-wise pose information. This network is also guided by view direction vectors $\mathcal{V}$ to model the view-dependent variation. We adopt the StyleUNet~\cite{wang2023styleavatar} architecture for this Gaussian U-Net:
\begin{equation}
    \Delta \boldsymbol{t}, \scale, \rot, \opacity, \col \leftarrow \mathcal{G}\left(\mathcal{P}(\boldsymbol{\Theta}), \mathcal{T}, \mathcal{V}\right)
\end{equation}
The network is designed to produce multiple Gaussian textures that encompass all the parameters required for 3D Gaussians. These parameters include the position residual $\Delta \boldsymbol{t}$, scale $\scale$, rotation $\rot$, opacity $\opacity$, and color $\col$. Notably, the position residual $\Delta \boldsymbol{t}$ plays a critical role in optimizing Gaussian point positions further based on the corresponding point positions in the refined mesh.

\subsection{Mesh-guided 3D Gaussians Animation}
\label{sec:animation}
Using a UV mask to filter out irrelevant texels in the texture maps, the remaining texels are transformed into Gaussian points in 3D space by UV mapping, resulting in approximately 190k Gaussian points with a texture map at $512 \times 512$ resolution. The final position of the Gaussian point $\mean$ is calculated by adding the rendered position map of the refined mesh and the position residual $\Delta \boldsymbol{t}$ of the Gaussian point. This position is combined with other parameters to produce the final image through rendering, utilizing a differentiable Gaussian rasterization.

\subsection{Training}
\label{sec:loss}
We jointly optimize the Mesh U-Net and Gaussian U-Net using multiple losses in our approach. When it comes to pose-dependent mesh deformation optimization, we utilize a per-frame SMPLX-D model to supervise the process. This mesh loss $\mathcal{L}_{mesh}$ includes a reconstruction term and a Laplacian term, which aims to guarantee that the optimized mesh not only conforms to the ground truth but also maintains smoothness:
\begin{equation}
    \mathcal{L}_{mesh} = \lambda_{recon}||\boldsymbol{G}_{gt} - \boldsymbol{G}_{refined}||_{2}^{2} + \lambda_{laplacian}\sum_{p}||\delta_{p}^{gt} - \delta_{p}^{refined}||_{2}^{2}
\end{equation}
The Laplacian regularization term is defined as the square of the $L_{2}$ norm of the difference between the ground truth Laplacian coordinate and the refined Laplacian coordinate. Here, the Laplacian coordinates for each vertex $p$ is calculated as the difference between the average coordinate of its neighbors ($k \in \mathcal{N}(p)$) and the vertex itself: $\delta_p = p - \frac{1}{\|\mathcal{N}(p)\|} \sum_{k \in \mathcal{N}(p)} k$. 

To supervise the rendered image, we incorporate L1 loss and SSIM loss~\cite{kerbl20233d}, as well as perceptual loss between rendered images and ground truth images to improve the semantic quality of the image.
\begin{equation}
    \mathcal{L}_{image} = \lambda_{l1}\mathcal{L}_{l1} + \lambda_{ssim}\mathcal{L}_{ssim} + \lambda_{perceptual}\mathcal{L}_{perceptual}
\end{equation}
Additionally, we introduce a regularization loss to constrain Gaussian residuals and scale.
\begin{equation}
    \mathcal{L}_{gaussian} = \lambda_{scale}||exp(\scale)||_{2}^{2} + \lambda_{res}||\Delta \boldsymbol{t}||_{2}^{2}
\end{equation}
This loss term not only expedites convergence but also enhances robustness. In summary, the overall loss is:
\begin{equation}
    \mathcal{L}=\mathcal{L}_{mesh} + \mathcal{L}_{image} + \mathcal{L}_{gaussian}
\end{equation}
The loss weights are set as: $\lambda_{recon} = 100, \lambda_{laplacian} = 500, \lambda_{l1} = 1, \lambda_{ssim} = 0.2, \lambda_{perceptual} = 0.2, \lambda_{scale} = 1, \lambda_{res} = 1$.

%% file: sec/4_experiments.tex
\section{Experiments}
In this section, we present the rendering results of drivable human avatars reconstructed using our method. We first give some details about the experiments. Then we provide quantitative and qualitative comparisons with several state-of-the-art algorithms~\cite{peng2021neural, peng2021animatable, chen2023uv, qian20233dgs}, showcasing the performance under novel view and novel pose tasks. Lastly, we conduct thorough ablation experiments to validate the effectiveness of our core modules.

\subsection{Experiment Setup}
\label{sec:ExperimentSetup}

\noindent\textbf{Implementation Details.} 
We implement the Mesh U-Net as a standard U-Net with 5 downsampling layers. And the Gaussian U-Net is a StyleUNet~\cite{wang2023styleavatar} constructed based on StyleGAN~\cite{karras2019stylegan} since we find StyleUNet structure captures the texture details better than the standard U-Net. Both models operate on texture space at a resolution of $512\times512$, encompassing position maps, average texture images, offset maps, and Gaussian textures. During training, we use the Adam~\cite{kingma2014adam} optimizer, applying learning rates of $2\times10^{-5}$ for Mesh U-Net and $2\times10^{-3}$ for Gaussian U-Net. The training is conducted with a batch size of 1, lasting for 1.5 million iterations, which requires approximately 3 days to complete on a single A100 GPU.

\noindent\textbf{Dataset.} 
To ensure the dense and uniform elevation and surround angles, we construct a $360^\circ$ camera array employing a total of 72 cameras for capturing the data. This includes $8$ 4K cameras primarily capturing front views (at $4096\times3072$ resolution) and $64$ 2K cameras (at $2592\times2048$ resolution). All cameras capture synchronized frames at 30 fps. To construct our dataset, we evenly select 64 cameras as training views and the remaining 8 cameras as testing views. We collect and process 5 motion videos with different subjects, fitting them frame by frame with the SMPL-X model, SMPLX-D model, and corresponding texture maps. We capture 900 frames for each sequence and select 400 consecutive frames from the middle of the sequence to form the dataset, discarding the relatively static frames at the beginning and end of the sequence. We use the first 300 frames as the training set and the last 100 frames as the testing set.


\noindent\textbf{Metric.} \ 
To assess image quality effectively, we utilize Peak Signal-to-Noise Ratio (PSNR), Structural Similarity Index Measure (SSIM), and Learned Perceptual Image Patch Similarity (LPIPS)~\cite{zhang2018unreasonable} metrics in quantitative experiments. 

\subsection{Comparison}

\begin{table}[tb]
\caption{Quantitative comparison of \textbf{novel view synthesis}. We highlight the best numbers in \textbf{bold} and the second best in \underline{underline}.}
\label{tab:comparison of novel view}
\centering
\resizebox{\textwidth}{!}{%
    \begin{tabular}{lccccccccccccccc}
    \toprule
    Subject & \multicolumn{3}{c}{S1} & \multicolumn{3}{c}{S2} & \multicolumn{3}{c}{S3} & \multicolumn{3}{c}{S4} & \multicolumn{3}{c}{S5} \\
    Metric & PSNR$\uparrow$ & SSIM$\uparrow$ & \multicolumn{1}{c|}{LPIPS$\downarrow$} & PSNR$\uparrow$ & SSIM$\uparrow$ & \multicolumn{1}{c|}{LPIPS$\downarrow$} & PSNR$\uparrow$ & SSIM$\uparrow$ & \multicolumn{1}{c|}{LPIPS$\downarrow$} & PSNR$\uparrow$ & SSIM$\uparrow$ & \multicolumn{1}{c|}{LPIPS$\downarrow$} & PSNR$\uparrow$ & SSIM$\uparrow$ & LPIPS$\downarrow$ \\ 
    \midrule
    \multicolumn{1}{l|}{NeuralBody\cite{peng2021neural}} & 26.26 & 0.9336 & \multicolumn{1}{c|}{0.1503} & 26.56 & 0.9424 & \multicolumn{1}{c|}{0.1417} & 25.37 & 0.9228 & \multicolumn{1}{c|}{0.1533} & 26.59 & 0.9363 & \multicolumn{1}{c|}{0.1522} & 26.10 & 0.9437 & 0.1307 \\
    \multicolumn{1}{l|}{Anim-NeRF\cite{peng2021animatable}} & 22.99 & 0.8946 & \multicolumn{1}{c|}{0.1942} & 24.43 & 0.9203 & \multicolumn{1}{c|}{0.1665} & 23.08 & 0.8959 & \multicolumn{1}{c|}{0.1918} & 25.04 & 0.9065 & \multicolumn{1}{c|}{0.1861} & 24.38 & 0.9231 & 0.1649 \\
    \multicolumn{1}{l|}{UV~Volumes\cite{chen2023uv}} & 23.47 & 0.9148 & \multicolumn{1}{c|}{0.1260} & 24.38 & 0.9297 & \multicolumn{1}{c|}{0.1043} & 23.06 & 0.9092 & \multicolumn{1}{c|}{0.1151} & 24.25 & 0.9168 & \multicolumn{1}{c|}{0.1143} & 24.36 & 0.9334 & 0.0986 \\
    \multicolumn{1}{l|}{3DGS-Avatar\cite{qian20233dgs}} & \underline{26.28} & \underline{0.9371} & \multicolumn{1}{c|}{\underline{0.1100}} & \underline{27.36} & \underline{0.9515} & \multicolumn{1}{c|}{\underline{0.0877}} & \underline{25.69} & \underline{0.9333} & \multicolumn{1}{c|}{\underline{0.1013}} & \underline{28.30} & \underline{0.9505} & \multicolumn{1}{c|}{\underline{0.0931}} & \underline{27.27} & \underline{0.9541} & \underline{0.0892} \\ 
    \midrule
    \multicolumn{1}{l|}{Ours} & \textbf{27.42} & \textbf{0.9557} & \multicolumn{1}{c|}{\textbf{0.0592}} & \textbf{28.42} & \textbf{0.9665} & \multicolumn{1}{c|}{\textbf{0.0471}} & \textbf{26.47} & \textbf{0.9554} & \multicolumn{1}{c|}{\textbf{0.0547}} & \textbf{28.90} & \textbf{0.9625} & \multicolumn{1}{c|}{\textbf{0.0492}} & \textbf{27.65} & \textbf{0.9655} & \textbf{0.0478} \\ 
    \bottomrule
    \end{tabular}%
}
\end{table}

\begin{table}[tb]
\caption{Quantitative comparison of \textbf{novel pose synthesis}.}
\label{tab:comparison of novel pose}
\centering
\resizebox{\textwidth}{!}{%
    \begin{tabular}{lccccccccccccccc}
    \toprule
    Subject & \multicolumn{3}{c}{S1} & \multicolumn{3}{c}{S2} & \multicolumn{3}{c}{S3} & \multicolumn{3}{c}{S4} & \multicolumn{3}{c}{S5} \\
    Metric & PSNR$\uparrow$ & SSIM$\uparrow$ & \multicolumn{1}{c|}{LPIPS$\downarrow$} & PSNR$\uparrow$ & SSIM$\uparrow$ & \multicolumn{1}{c|}{LPIPS$\downarrow$} & PSNR$\uparrow$ & SSIM$\uparrow$ & \multicolumn{1}{c|}{LPIPS$\downarrow$} & PSNR$\uparrow$ & SSIM$\uparrow$ & \multicolumn{1}{c|}{LPIPS$\downarrow$} & PSNR$\uparrow$ & SSIM$\uparrow$ & LPIPS$\downarrow$ \\ 
    \midrule
    \multicolumn{1}{l|}{Neural~Body\cite{peng2021neural}} & {\ul 23.77} & {\ul 0.9182} & \multicolumn{1}{c|}{0.1517} & {\ul 23.68} & {\ul 0.9197} & \multicolumn{1}{c|}{0.1507} & {\ul 21.03} & {\ul 0.889} & \multicolumn{1}{c|}{0.1742} & 23.69 & \textbf{0.9188} & \multicolumn{1}{c|}{0.1568} & 21.38 & 0.9088 & 0.1522 \\
    \multicolumn{1}{l|}{Anim-NeRF\cite{peng2021animatable}} & 22.39 & 0.8911 & \multicolumn{1}{c|}{0.1897} & 23.16 & 0.9098 & \multicolumn{1}{c|}{0.1683} & 20.24 & 0.8675 & \multicolumn{1}{c|}{0.2092} & \textbf{24.22} & 0.9073 & \multicolumn{1}{c|}{0.1753} & {\ul 22.05} & 0.9066 & 0.1707 \\
    \multicolumn{1}{l|}{UV~Volumes\cite{chen2023uv}} & 22.37 & 0.9064 & \multicolumn{1}{c|}{0.1325} & 22.57 & 0.9095 & \multicolumn{1}{c|}{0.1239} & 20.24 & 0.8782 & \multicolumn{1}{c|}{0.1514} & 23.02 & 0.9124 & \multicolumn{1}{c|}{0.1210} & 20.10 & 0.8969 & 0.1409 \\
    \multicolumn{1}{l|}{3DGS-Avatar\cite{qian20233dgs}} & 22.48 & 0.9084 & \multicolumn{1}{c|}{{\ul 0.1221}} & 23.07 & 0.9166 & \multicolumn{1}{c|}{{\ul 0.1060}} & 20.56 & 0.8812 & \multicolumn{1}{c|}{{\ul 0.1351}} & 23.08 & 0.9130 & \multicolumn{1}{c|}{{\ul 0.1145}} & 21.34 & {\ul 0.9119} & {\ul 0.1234} \\ 
    \midrule
    \multicolumn{1}{l|}{Ours} & \textbf{24.41} & \textbf{0.9307} & \multicolumn{1}{c|}{\textbf{0.0857}} & \textbf{23.80} & \textbf{0.9230} & \multicolumn{1}{c|}{\textbf{0.0967}} & \textbf{21.84} & \textbf{0.8978} & \multicolumn{1}{c|}{\textbf{0.1158}} & {\ul 24.20} & \textbf{0.9188} & \multicolumn{1}{c|}{\textbf{0.1060}} & \textbf{22.77} & \textbf{0.9224} & \textbf{0.1106} \\ 
    \bottomrule
    \end{tabular}%
}
\end{table}

\begin{figure}[htbp]
	\centering
	\includegraphics[width=1.0\columnwidth]{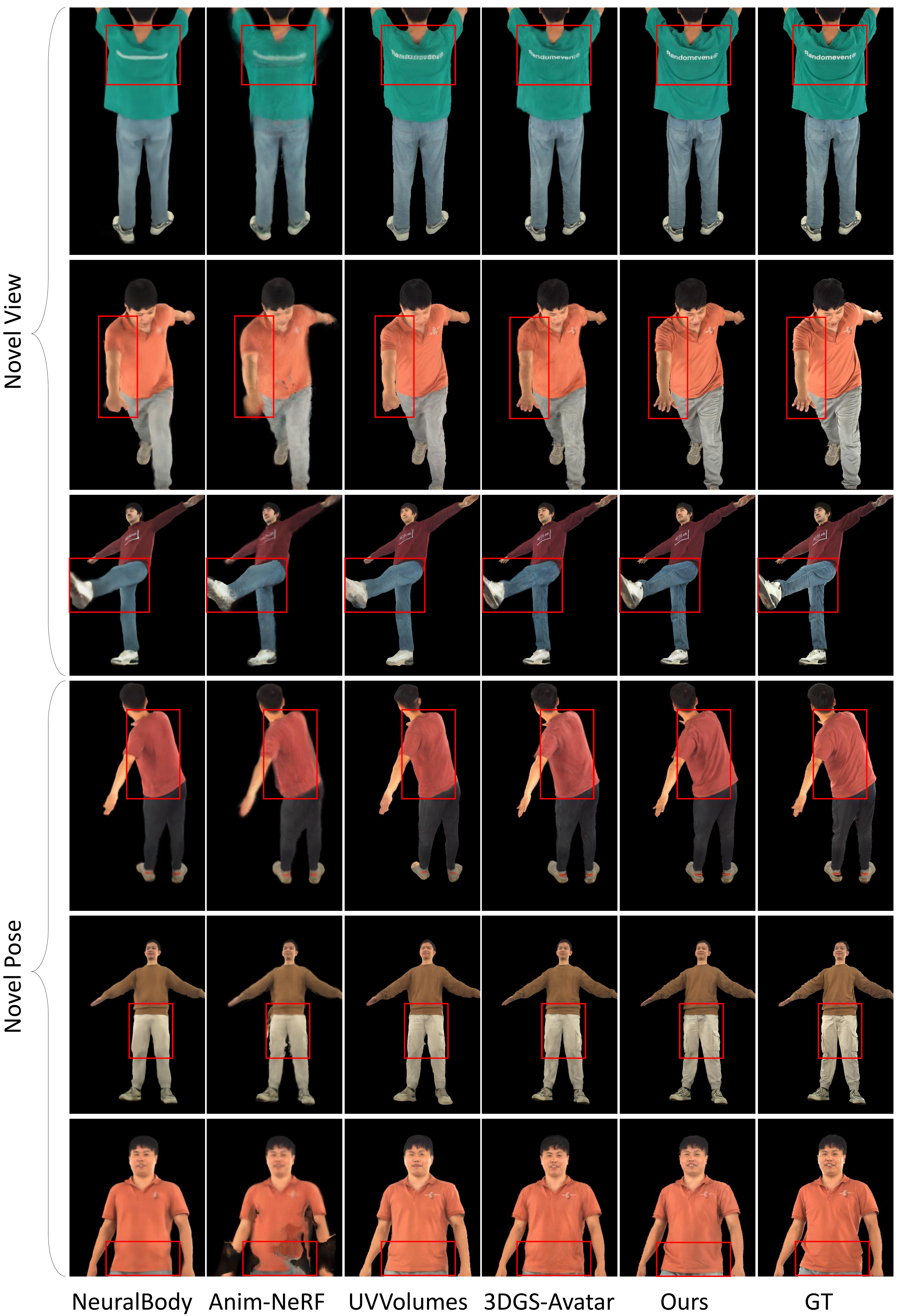}
	\caption{Qualitative comparisons on novel view and novel pose synthesis.}
        \label{fig:comparison}
\end{figure}

We compare our algorithm with several baselines: Neural Body~\cite{peng2021neural}, Anim-NeRF~\cite{peng2021animatable}, UV Volumes~\cite{chen2023uv}, and a recent state-of-the-art human avatar methods using GS with source code available, 3DGS-Avatar~\cite{qian20233dgs}. These methods also utilize the SMPL model as their input; hence, we also apply the SMPL model fitting procedure to each frame in the dataset. Additionally, we also process some supplementary data following the guidelines of the official code. Specifically, we handle the LBS weights for Anim-NeRF, perform DensePose~\cite{guler2018densepose} estimation for UV Volumes, and reprocess the SMPL parameters for 3DGS-Avatar. It is worth noting that 3DGS-Avatar is originally designed for monocular video input. We expand it to include multi-view input under the same settings for a fair comparison. The complete experimental setup can be found in the supplemental materials~\ref{sec:supp-exp details}.

The quantitative results for novel view synthesis can be seen in Table~\ref{tab:comparison of novel view}, while the novel pose synthesis results are presented in Table~\ref{tab:comparison of novel pose}. Our algorithm achieves the best performance across all metrics in the novel view synthesis experiment. For novel pose synthesis, we outperform other baselines for most subjects, except S4, where we achieve comparable results with Anim-NeRF. Note that Anim-NeRF requires an additional training stage for rendering novel poses, which introduces computational overhead to perform the task.
The qualitative results for novel view and novel pose synthesis are presented in Fig.~\ref{fig:comparison}. Our method can render photorealistic novel view images by preserving numerous clothing wrinkles and texture details. In contrast, while 3DGS-Avatar recovers some deformations, its rendering results still exhibit significant noise and detail loss. The other three NeRF-based methods produce prominent artifacts and blurriness in challenging poses, particularly struggling with intricate regions like fingers. 
 For more detailed image results and character-driven video outcomes, readers are encouraged to refer to the supplementary materials~\ref{sec:supp-additional results}.

\begin{table}[tb]
\caption{Quantitative results of ablation study testing on novel view and novel pose synthesis.}
\label{tab:ablation}
\centering

\resizebox{\textwidth}{!}{%
    \begin{tabular}{lcccccc}
    \toprule
     & \multicolumn{3}{c}{Novel View} & \multicolumn{3}{c}{Novel Pose} \\
    Metric & PSNR$\uparrow$ & SSIM$\uparrow$ & \multicolumn{1}{c|}{LPIPS$\downarrow$} & PSNR$\uparrow$ & SSIM$\uparrow$ & LPIPS$\downarrow$ \\ \midrule
    w/o Mesh & 24.49 & 0.9202 & \multicolumn{1}{c|}{0.1577} & 17.99 & 0.8574 & 0.2051 \\
    w/ SMPL Mesh & \textbf{27.33} & \textbf{0.9548} & \multicolumn{1}{c|}{0.0619} & 21.43 & 0.8975 & 0.1326 \\
    w/ SMPLX-D Mesh & 27.26 & 0.9544 & \multicolumn{1}{c|}{0.0609} & 21.84 & 0.9009 & 0.1331 \\
    Full Model & 27.25 & 0.9544 & \multicolumn{1}{c|}{\textbf{0.0599}} & \textbf{21.88} & \textbf{0.9011} & \textbf{0.1295} \\
    \bottomrule
    \end{tabular}%
}
\end{table}

\begin{figure}[tb]
	\centering
	\includegraphics[width=1.0\columnwidth]{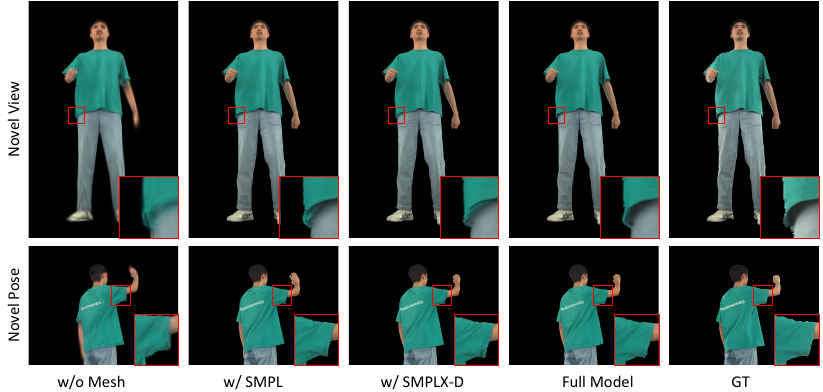}
	\caption{Novel view and pose synthesis results of the ablation experiments. The full model exhibits the least artifacts compared with using different mesh guidance or not using any mesh.}
        \label{fig:ablation}
\end{figure}
 
\subsection{Ablation Study}

In this section, we perform ablation experiments on the essential modules and loss terms employed in our method to validate the effectiveness of the design. For each subject, we use the same training/test split as in the comparison experiments. We assess the quality of novel pose synthesis using the subsequent 100 frames.

\noindent\textbf{Mesh Guidance.}
In the pipeline, we employ a mesh U-Net to predict pose-dependent vertex deformation, yielding a detailed mesh to guide the Gaussians. We validate this design by conducting three ablation experiments. \textit{w/o Mesh}: we directly predict the absolute coordinates of Gaussians without using any mesh guidance; \textit{w/ SMPL}: we use SMPL mesh as guidance and do not predict vertex offset; \textit{w/ SMPLX-D}: similar to \textit{w/ SMPL} but use SMPLX-D mesh without mesh U-Net as guidance. 

The quantitative results on novel view and novel pose synthesis are shown in Tab.~\ref{tab:ablation} and visual results are presented in Fig.~\ref{fig:ablation}. As observed, without utilizing mesh guidance, the learned Gaussians struggle with missing body parts and texture details, resulting in a significant decline across all metrics. The use of SMPL and SMPLX-D alleviates the artifacts and gets overall promising results. However, they may produce artifacts in regions with complex geometry where these two meshes fail to model accurately as shown in Fig.~\ref{fig:ablation}. Moreover, the estimation of SMPL may fail in regions like hands which could cause an error in the rendered image as shown in the bottom example of Fig.~\ref{fig:ablation}. Our full model restores better geometric details and reduces the occurrence of these artifacts. It is important to note that, due to differences typically found in local regions and texture details, the numerical results in Table~\ref{tab:ablation} may not exhibit significant improvements. Nonetheless, our full model still gets competitive numerical results in novel view synthesis and better results in novel pose synthesis.

Please refer to the supplementary materials~\ref{sec:supp-ablation study} for the loss ablation experiments.



%% file: sec/5_conclusion.tex
\section{Conclusion}
In this paper, we introduce a method called UV Gaussians, which combines 3D Gaussians and UV space representation. This approach enables the reconstruction of photo-realistic, pose-driven avatar models from multi-view images. Our method takes as input the position map of the model vertices, learns pose-dependent geometric deformations through Mesh U-Net, and learns the properties of Gaussian points embedded in UV space through Gaussian U-Net. Subsequently, guided by the refined mesh, rendering of Gaussian points is conducted to obtain rendered images from arbitrary viewpoints. By incorporating fine-grained geometric guidance and leveraging the feature learning capability of powerful 2D networks in UV space, our method achieves state-of-the-art results in experiments on novel view and novel pose synthesis. 

\noindent\textbf{Limitations.} \ 
Despite its achievements, our method is constrained by the dependence on scanned mesh. While minor fitting errors can be optimized using Mesh U-Net, significant errors may impact the performance of the method. Moreover, the dataset we collected does not include extremely loose clothing such as long skirts. In future research, we plan to evaluate our method on more available datasets containing multi-view images and scanned models, especially exploring challenging clothing types with difficult poses.

%% file: sec/6_suppl.tex
\section{Details of Data Processing}
\label{sec:supp-dataprocess}

\begin{figure}[htbp]
	\centering
	\includegraphics[width=1.0\columnwidth]{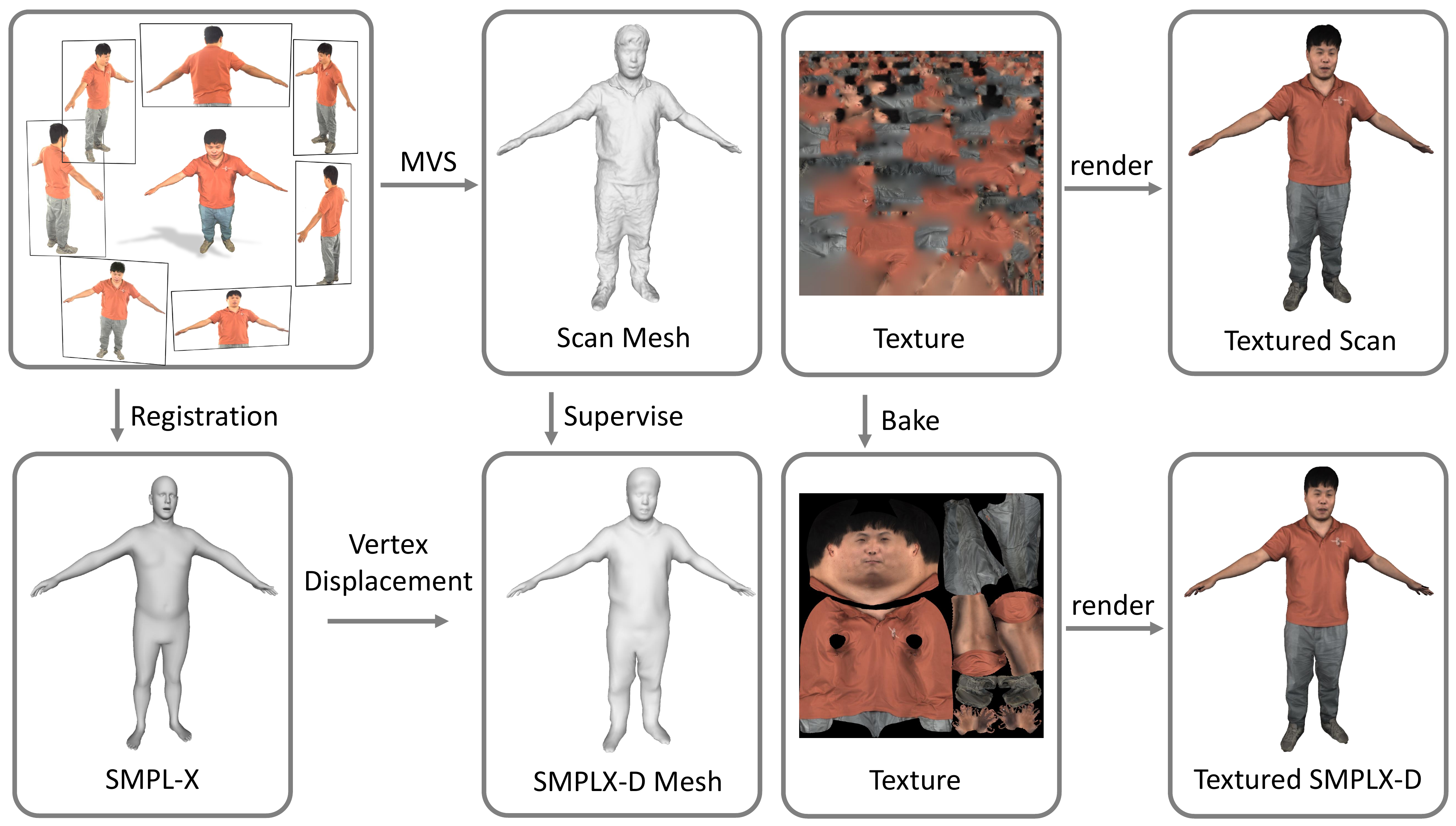}
	\caption{\textbf{The workflow of our data processing approach.}}
    \label{fig:dataprocess}
\end{figure}

We begin the synchronous capture using a total number of 72 cameras, comprising 8 4K cameras facing forward and 64 2K surround cameras, as discussed in Sec.~\ref{sec:ExperimentSetup} of the main paper. Actors participating in our dataset are well-informed and acknowledge that the data will be made public for research purposes.
The complete processing workflow is presented in Fig.~\ref{fig:dataprocess}. Utilizing the captured multi-view images and calibrated camera parameters, we employ EasyMocap~\cite{easymocap} to fit the SMPL~\cite{SMPL:2015} and SMPL-X~\cite{SMPL-X:2019} models. To enhance the parametric model's capability to capture clothing wrinkles and texture details effectively, we reconstruct the scanned model for each frame through the MVS method and follow the work of ~\cite{ma2020learning, bhatnagar2020combining} to fit the SMPLX-D model. Due to the increased topological complexity of the SMPL-X model in regions such as the eyes and lips, vertex and face reduction and smoothing operations are necessary when incorporating vertex displacement to fit the SMPLX-D model.

In addition to geometric fitting, we also conduct texture map fitting. The texture maps generated by the MVS method exhibit a relatively random topology, which needs to be adjusted to a uniform topology. We utilize the UV mapping included in the SMPL-X~\cite{SMPL-X:2019} model and modify it based on vertex-face processing of the SMPLX-D model to create a unified UV mapping. Texture transfer is achieved using the bake function in Blender software. To obtain the averaged texture map for each character's data, we average the texture maps of all frames, which serves as one of the inputs for our approach.

\section{Comparison}
\label{sec:supp-comparison}
In this section, we provide a comprehensive description of the implementation details for the baselines used in the comparative experiments and present additional qualitative results of comparisons.
During our study, we find that the multi-view approaches utilizing Gaussian Splatting are not available as open-source. Therefore, we choose to extend the single-view method 3DGS-Avatar~\cite{qian20233dgs} into a multi-view training approach.

\subsection{Experimental Details}
\label{sec:supp-exp details}
During the training phase, our method uses image resolutions of $648\times 512$ and $1024\times 768$, representing $\times 4$ down-sampling of 2K and 4K captured images, with 64 training viewpoints and 8 testing viewpoints. The training dataset consists of 300 frames, while there are 100 frames for testing novel poses. The baselines follow the same training-testing split and down-sampling factors. To ensure code compatibility, we fit both the SMPL-X and SMPL models simultaneously for the baselines.

To maintain a consistent evaluation standard, for all test frames, we save the corresponding $bound\_mask$ to evaluate different methods. Calculations for the metrics are confined to the pixels within the $bound\_mask$. The specific adjustments implemented for each of the distinct baselines are detailed in the following discussion.


\noindent\textbf{Neural Body.}\ 
We utilize the official code repository\footnote{https://github.com/zju3dv/neuralbody} to train models on our dataset and include evaluation code for the LPIPS metric.

\noindent\textbf{Anim-NeRF.}\ 
We utilize the official code repository\footnote{https://github.com/zju3dv/animatable\_nerf} for data processing and model training. Initially, we employ the official $prepare\_blend\_weights$ code to obtain the necessary training weights related to LBS. Subsequently, we train the model using the same experimental setup and also include the LPIPS metric in the evaluation process.

\noindent\textbf{UV Volumes.}\ 
Firstly, we process all images using DensePose\footnote{https://github.com/facebookresearch/detectron2/tree/main/projects/DensePose} following the instructions from the official code repository\footnote{https://github.com/fanegg/UV-Volumes}. Subsequently, we train the model using the official code and incorporate testing on novel poses. Finally, We ensure standardized evaluation using VGG for the LPIPS assessment.

\noindent\textbf{3DGS-Avatar.}\ 
First, we process and save the necessary SMPL parameters and images for training using the code from the ARAH repository\footnote{https://github.com/taconite/arah-release} according to official guidelines. Following this, we train the model using the official code repository\footnote{https://github.com/mikeqzy/3dgs-avatar-release} and incorporate novel pose testing code. It is worth noting that the original code conduct training with a single viewpoint. To maintain consistency in the experimental setup, we switch to training with multiple viewpoints.

\subsection{Additional Qualitative Results}
\label{sec:supp-additional results}
\begin{figure}[htbp]
	\centering
	\includegraphics[width=1.0\columnwidth]{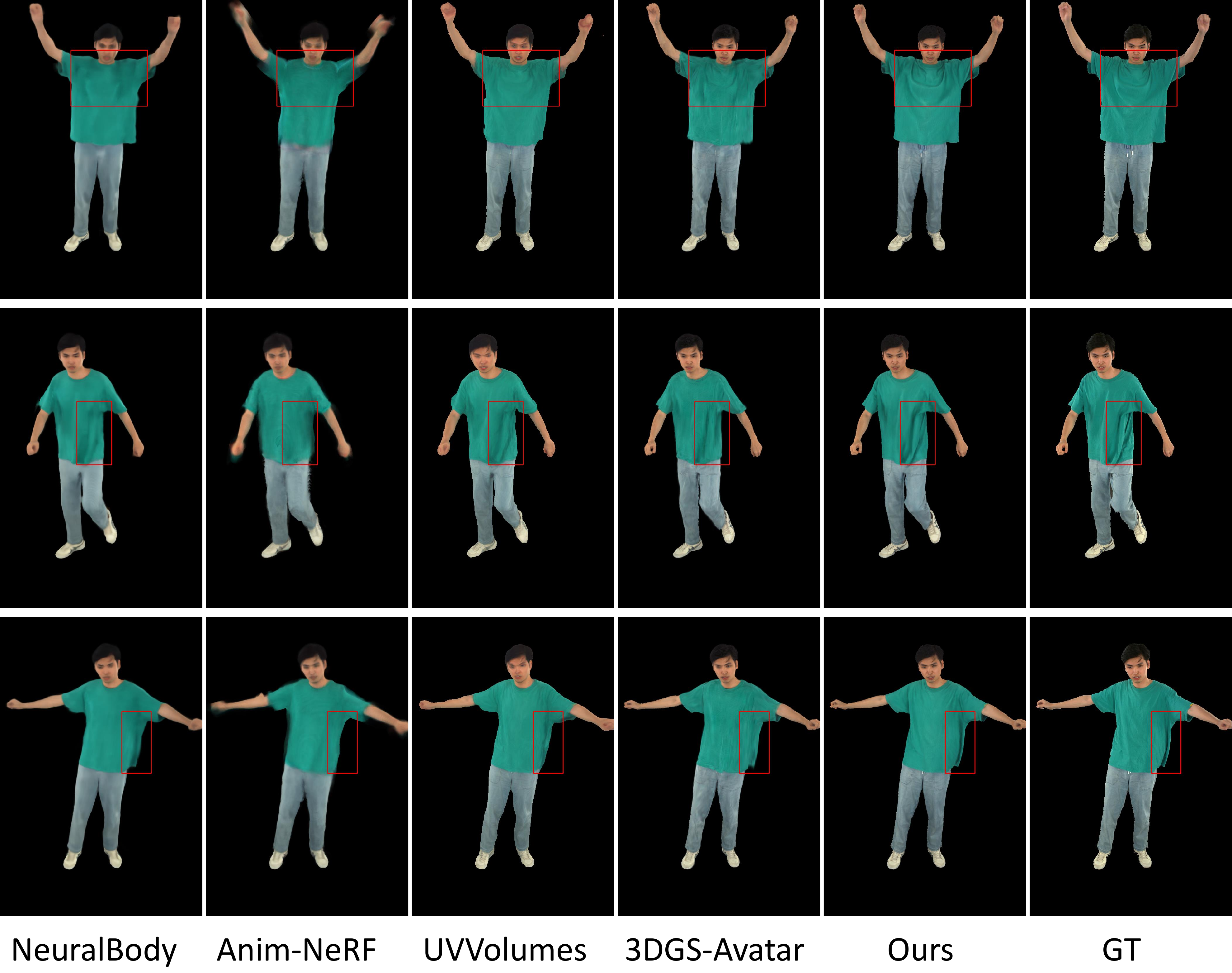}
	\caption{\textbf{Additional Comparison Results with Baselines.} Our method can preserve more clothing folds and texture details, resulting in clearer renderings.}
    \label{fig:additional comparison}
\end{figure}

In Fig.~\ref{fig:additional comparison}, we present additional qualitative comparison results. 
Our method can render photorealistic novel view images by preserving numerous clothing wrinkles and texture details. 
\textbf{For enhanced visualization, we highly recommend checking out our supplementary video.}

\section{Ablation Study}
\label{sec:supp-ablation study}
In this section, we conduct additional ablation experiments to verify the influence of these designs.

\subsection{Loss Function}

\begin{figure}[htbp]
	\centering
	\includegraphics[width=1.0\columnwidth]{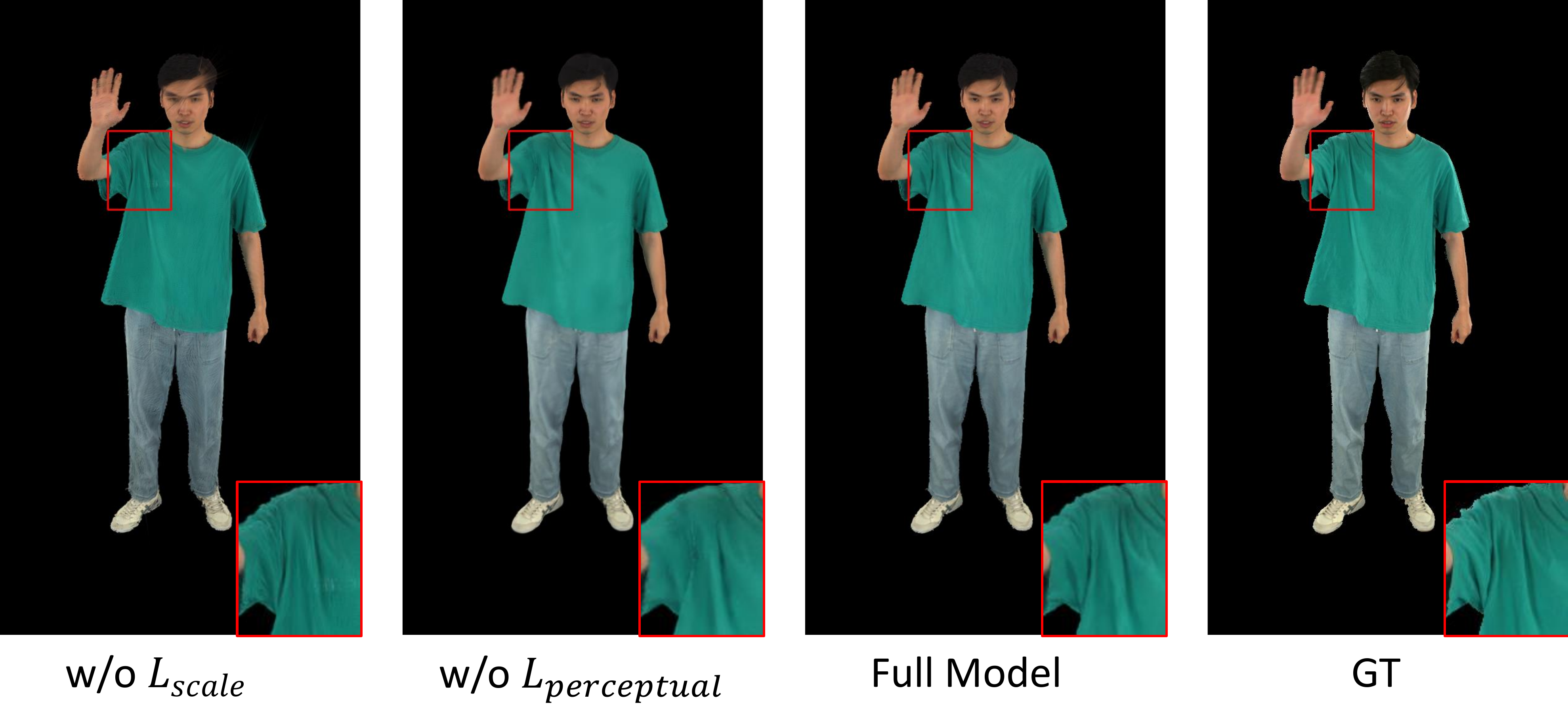}
	\caption{\textbf{Ablation Study of Loss.} The lack of $\mathcal{L}_{scale}$ causes notable artifacts in the head region. Without $\mathcal{L}_{perceptual}$, the output becomes more blurred, resulting in a loss of high-frequency details. Our full model can achieve better rendering effects.}
    \label{fig:loss ablation}
\end{figure}

\begin{table}[htbp]
\caption{\textbf{Ablation Study on Loss Terms}. Please note that experiments training without $\mathcal{L}_{res}$ or $\mathcal{L}_{mesh}$ do not converge.}
\label{tab:loss ablation}
\centering
\resizebox{\textwidth}{!}{%
    \begin{tabular}{lcccccc}
    \toprule
     & \multicolumn{3}{c}{Novel View} & \multicolumn{3}{c}{Novel Pose} \\
    Metric & PSNR$\uparrow$ & SSIM$\uparrow$ & \multicolumn{1}{c|}{LPIPS$\downarrow$} & PSNR$\uparrow$ & SSIM$\uparrow$ & LPIPS$\downarrow$ \\
    \midrule
    w/o $\mathcal{L}_{res}$ & - & - & \multicolumn{1}{c|}{-} & - & - & - \\
    w/o $\mathcal{L}_{mesh}$ & - & - & \multicolumn{1}{c|}{-} & - & - & - \\
    w/o $\mathcal{L}_{scale}$ & 26.11 & 0.9423 & \multicolumn{1}{c|}{\ul 0.1017} & 21.58 & 0.8919 & {\ul 0.151} \\
    w/o $\mathcal{L}_{perceptual}$ & \textbf{27.75} & \textbf{0.9558} & \multicolumn{1}{c|}{0.104} & \textbf{22.07} & {\ul 0.9008} & 0.1545 \\
    Full Model & {\ul 27.25} & {\ul 0.9544} & \multicolumn{1}{c|}{\textbf{0.0599}} & {\ul 21.88} & \textbf{0.9011} & \textbf{0.1295} \\
    \bottomrule
    \end{tabular}%
}
\end{table}

\begin{table}[htbp]
\caption{\textbf{Ablation Study of Training Frames.}}
\label{tab:frames ablation}
\centering
\resizebox{\textwidth}{!}{%
    \begin{tabular}{lcccccc}
    \toprule
     & \multicolumn{3}{c}{Novel View} & \multicolumn{3}{c}{Novel Pose} \\
     & PSNR$\uparrow$ & SSIM$\uparrow$ & \multicolumn{1}{c|}{LPIPS$\downarrow$} & PSNR$\uparrow$ & SSIM$\uparrow$ & LPIPS$\downarrow$ \\ 
    \midrule
    50 & 27.39 & 0.9553 & \multicolumn{1}{c|}{0.0604} & 21.96 & 0.9027 & 0.1275 \\
    100 & 27.47 & 0.9561 & \multicolumn{1}{c|}{\textbf{0.0582}} & 22.23 & 0.9058 & 0.1188 \\
    200 & {\ul 27.53} & {\ul 0.9564} & \multicolumn{1}{c|}{{\ul 0.0585}} & {\ul 23.13} & {\ul 0.9166} & {\ul 0.1033} \\
    300 & \textbf{27.57} & \textbf{0.9567} & \multicolumn{1}{c|}{0.059} & \textbf{24.41} & \textbf{0.9307} & \textbf{0.0857} \\ 
    \bottomrule
    \end{tabular}%
}
\end{table}

In Sec.~\ref{sec:loss} of the main paper, we introduce the training loss terms we utilize in our method. We conduct ablation experiments on these terms, specifically $\mathcal{L}_{mesh}$, $\mathcal{L}_{perceptual}$, $\mathcal{L}_{scale}$, and $\mathcal{L}_{res}$, with the quantitative results presented in Tab.~\ref{tab:loss ablation}. 
Experiments training without $\mathcal{L}_{mesh}$ or $\mathcal{L}_{res}$ do not converge, so we only show the qualitative results of the $w/o\ \mathcal{L}_{perceptual}$ and $w/o\ \mathcal{L}_{scale}$ in the Fig.~\ref{fig:loss ablation}.

The LPIPS metric of the full model outperforms other experimental results significantly.
While the model without $\mathcal{L}_{perceptual}$ achieves a higher PSNR than the full model, the rendering results of the former appear more blurry in comparison, as they lack many high-frequency details. Furthermore, both the quantitative and qualitative outcomes of the model without $\mathcal{L}_{scale}$ are significantly inferior to those of the full model, resulting in noticeable artifacts in the head and multiple body regions.

\subsection{The Number of Training Frames}

\begin{figure}[htbp]
	\centering
	\includegraphics[width=1.0\columnwidth]{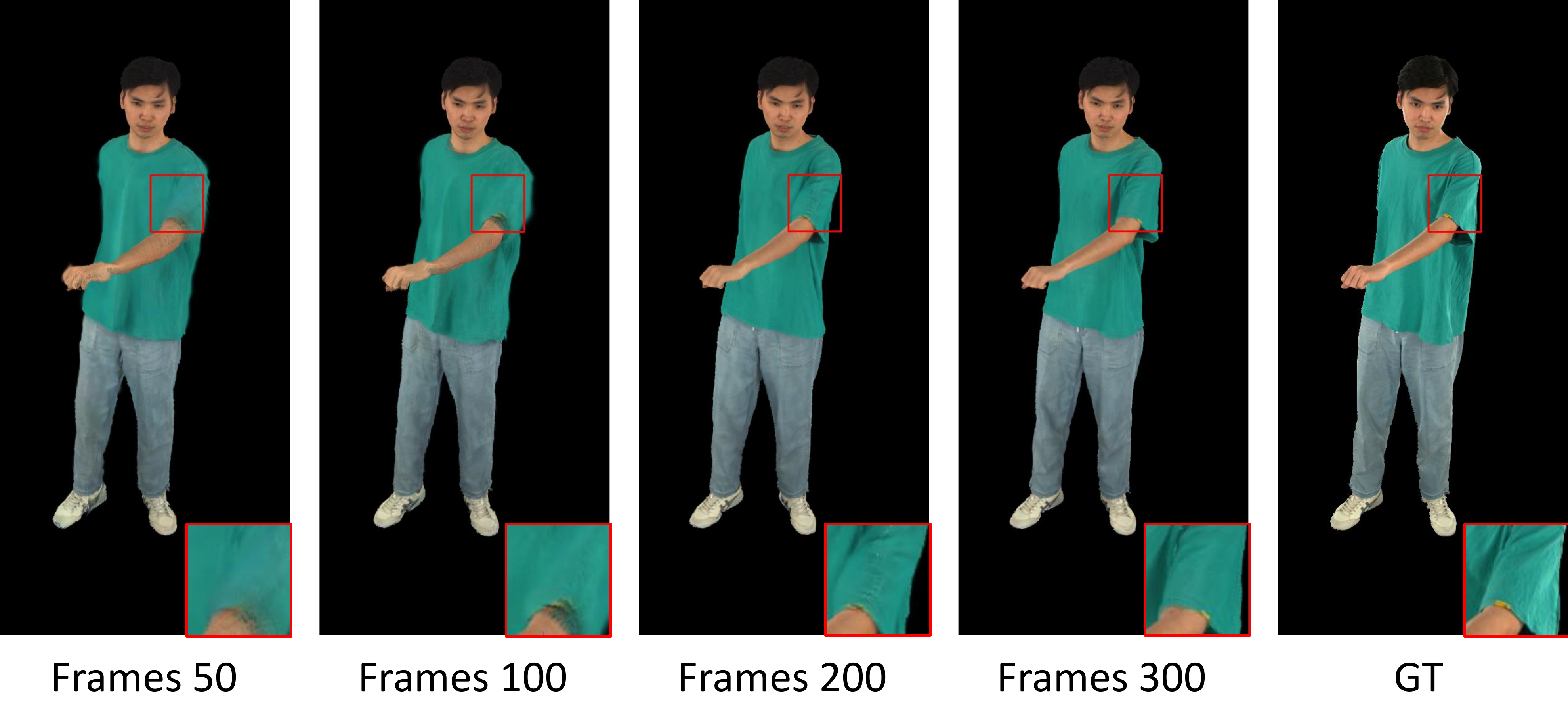}
	\caption{\textbf{Ablation Study on Training Frames.} Longer training frames achieve better generalization performance and demonstrate superior rendering effects on novel pose synthesis.}
    \label{fig:frames ablation}
\end{figure}

We conduct an ablation study using training frames of different lengths, as shown in the Fig.~\ref{fig:frames ablation} and Tab.~\ref{tab:frames ablation}. 
We perform novel view testing on frames used in training, and novel pose testing on test frames not used in any experiments. We can observe that longer training frames achieve better generalization performance and demonstrate superior rendering effects on novel pose synthesis.
In the area outlined in Fig.~\ref{fig:frames ablation}, the rendering results of the 50-frame model are very blurry with almost no details. As the number of training frames increases, the boundaries of the clothing and arms become clearer, accompanied by a reduction of artifacts. The 300-frame model retains texture details effectively, leading to the achievement of high photorealistic rendering quality.

%
%